\documentclass[11pt,en]{elegantpaper}

\usepackage[utf8]{inputenc} 
\usepackage[T1]{fontenc}    
\usepackage{hyperref}       
\usepackage{url}            
\usepackage{booktabs}       
\usepackage{amsfonts}       
\usepackage{nicefrac}       
\usepackage{microtype}      
\usepackage{lipsum}
\usepackage[autostyle]{csquotes} 
\usepackage[symbol, flushmargin]{footmisc}
\usepackage[final]{pdfpages}

\title{Trouble with the Curve: Predicting Future MLB Players Using Scouting Reports}

\author{
 Jacob Danovitch\thanks{Work primarily completed during internship at Microsoft.} \\
 Department of Computer Science\\
Carleton University\\
Ottawa, Canada \\
 \texttt{jacob.danovitch@carleton.ca} \\
}

\begin{document}
\maketitle

\begin{abstract}
In baseball, a scouting report profiles a player's characteristics and traits, usually intended for use in player valuation. This work presents a first-of-its-kind dataset of almost 10,000 scouting reports for minor league, international, and draft prospects. Compiled from articles posted to MLB.com and Fangraphs.com, each report consists of a written description of the player, numerical grades for several skills, and unique IDs to reference their profiles on popular resources like MLB.com, FanGraphs, and Baseball-Reference. 

With this dataset, we employ several deep neural networks to predict if minor league players will make the MLB given their scouting report. We open-source this data to share with the community, and present a web application demonstrating language variations in the reports of successful and unsuccessful prospects.

\end{abstract}

\section{Introduction}
In the 2012 film \textit{Trouble with the Curve}, Clint Eastwood portrays a scout for the Atlanta Braves whose traditional methods lie at odds with new statistical analyses. The film shares a look into the divide between the two major schools of thought in professional sports - the 'eye test', and 'advanced analytics.' The former method relies on closely studying players and making evaluations based on expert knowledge, whereas the latter advocates for data-driven decision making and the use of statistics to project future value. The work presented in this paper argues that this divide presents a false dichotomy - that both the eye test and advanced analytics have a place in the evaluation of baseball players.

To find a common ground between both sides, we use statistical and deep learning methods in combination with the expert knowledge contained in scouting reports to predict if an amateur baseball player will make it to the major leagues. Further, we demonstrate language variations in the reports of successful and unsuccessful prospects, and we open-source this data to share with the community\footnote[7]{\url{https://github.com/jacobdanovitch/Trouble-With-The-Curve}}. This dataset contains almost 10,000 scouting reports for minor league, international, and draft prospects, consisting of written descriptions, numerical grades, and various pieces of metadata.

\section{Related Work}
\label{sec:headings}

Projecting the future performance of amateur athletes has long been of interest to analysts of all sports. However, most previous contributions have focused on the use of player statistics as features; very few have attempted a text mining approach. As such, we believe this dataset to be the first open-source collection of baseball scouting reports, and the largest open-source collection of scouting reports for any sport.

\paragraph{Statistical projections.} Given the absence of scouting reports such as those presented here, most attempts to project the future performance of amateur baseball players have used box-score data and statistical measures as features. One such example is the KATOH system, developed to project the future performance of minor league baseball players. The system uses each player's statistics to model a probability distribution of future levels of performance, as well as estimating the probability that the player will make the major leagues. The author notes that the system fails to account for specific player traits like defense and speed, leading it to "underrate players who man premium defensive positions–like catcher, center field, and shortstop–whose offensive abilities may not be the most valuable part of their game" \cite{mitchellkatoh}. Similar work was presented in \cite{carruthersprimes}, which developed a method to translate minor league statistics to future performance. The method used Weighted Runs Created Plus (wRC+), a measure of offensive performance relative to league average, as well as the player's age and level of competition to project their future wRC+ in the major leagues. Lastly, in a three-part series \cite{ringerreports} that appeared in the popular sports publication 'The Ringer', the authors were able to acquire 73,000 scouting reports from a former member of the Cincinnati Reds baseball organization. Unfortunately, the authors did not examine the written descriptions, opting instead to use numeric grades (also presented in our dataset). These grades demonstrated a moderate correlation to both player performance and career length, though use of the written descriptions could have created a stronger result.

\paragraph{Text mining approaches.} Comparatively speaking, little work has been done using written descriptions to project future performance. \cite{seppatext} presents an excellent analysis of hockey scouting reports using text mining, performing keyword and topic extraction and modeling current player performance using written descriptions. Unfortunately, the dataset used in their work is not readily available. Their work was expanded upon in a later contribution very similar to that presented here. \cite{liuprospectclf} constructs a dataset of scouting reports for amateur hockey players and uses the reports to predict the success of each prospect, defined as whether a player appeared in more games than the average for his cohort. This work shares the same intent as ours, and is entirely open source. The two distinctions of our work are (1) the size of the dataset (just under 1300 reports, versus nearly 10,000 for ours) and (2) the sport in question (hockey versus baseball). Hopefully, future work will replicate these studies for a wide variety of other sports.

\section{The TWTC Dataset}

\subsection{Scouting Reports}

The main contribution of this work is to open-source the \textit{Trouble with the Curve} (TWTC) dataset for use by the community. The dataset contains almost 10,000 scouting reports for amateur baseball players of varying levels, such as minor leaguers, international prospects, and draft prospects. Each scouting report consists of several features.

\paragraph{Written descriptions.} Each report features a parapgraph-length description of the player's strengths and weaknesses. To the best of our knowledge at the time of writing, this is the first public dataset containing these descriptions. The descriptions are written by MLB.com Prospect Pipeline and Fangraphs.com writers, well-respected staffs who provide extensive coverage of amateur baseball. Each description will generally summarize the recent performance of the player, describe strengths and weaknesses, and project future performance. The descriptions use highly domain-specific language, as shown in the example below (emphasis added):

\blockquote{
    "Andujar received the highest bonus in the Yankees' 2011 international class, signing for \$750,000 out of Venezuela. As a 19-year-old in the low Class A South Atlantic League, he struggled early in 2014 but rebounded to hit .319/.367/.456 in the second half. Andujar combines bat speed with an advanced approach for his age. He has shown an ability to catch up to quality fastballs and make adjustments against offspeed pitches. He has the potential to produce 20 or more homers per season, and if he can develop some more plate discipline, he should hit for average as well. Though he has committed 51 errors in 196 pro games at third base, Andujar has the tools to become a capable defender. He has \emph{a cannon arm and good hands} but needs to learn to \emph{not try to do too much at the hot corner}."
}

\paragraph{Numeric grades.} Each report assigns numeric grades for a variety of skills to each player, such as batter's ability to hit for power or run quickly, the speed and control of a pitcher's fastball, and so on. Grades are assigned along the "20-80 scale", a fixture in baseball scouting. The player is assigned a grade between 20 and 80 for each skill, with 50 as the average score, and every 10 points representing one standard deviation from the mean. These grades are generally then aggregated into an overall score for the player as a rough estimate of overall value. 

Importantly, the grades are \textit{projections}, rather than evaluations; the goal of scouting is to predict future performance. A player with a grade of 20 is expected to be inadequate to consistently play at the major league level, and would only be called up in the case of injury; we call this player "replacement-level". A player with a grade of 80 is expected to be among the best in the league, potentially even a hall-of-fame candidate. Table \ref{tab:fg-scale}, the data for which was originally presented in \cite{fgscouting}, illustrates this by relating each score to an expected amount of Wins Above Replacement (WAR), an estimate of how many wins each player contributes to their team relative to a replacement-level baseline.

The TWTC dataset presents grades for: batting (contact hitting, power); defense (speed, fielding, throwing strength); and pitching (a grade for each pitch type, as well as overall control). 

\begin{table}[h]
\centering
\begin{tabular}{|l|l|l|}
\hline
\textbf{Scouting Grade} & \textbf{Role} & \textbf{WAR} \\ \hline
20 & Replacement level & – \\ \hline
30 & Organizational depth & < -0.1 \\ \hline
40 & Bench Player & 0.0 to 0.9 \\ \hline
45 & Low End Starter & 1.0 to 1.7 \\ \hline
50 & Average Starter & 1.8 to 2.5 \\ \hline
55 & Above Average Starter & 2.6 to 3.4 \\ \hline
60 & All Star & 3.5 to 4.9 \\ \hline
70 & Top 10 overall & 5.0 to 7.0 \\ \hline
80 & Top 5 overall & >7.0 \\ \hline
\end{tabular}
\caption{The expected role and value for prospects by grade. \cite{fgscouting}.}
\label{tab:fg-scale}
\end{table}

\paragraph{Identifiers.} Lastly, each report contains several unique identifier keys to link each profile to the player's statistics on popular resources such as MLB.com and FanGraphs.com. This presents an exciting opportunity for further studies on the predictive value of these scouting reports, such as modeling future WAR instead of a binary classification.

\subsection{Corpus Visualization}


\includegraphics[trim={7cm 0 0 0}, clip,scale=0.5]{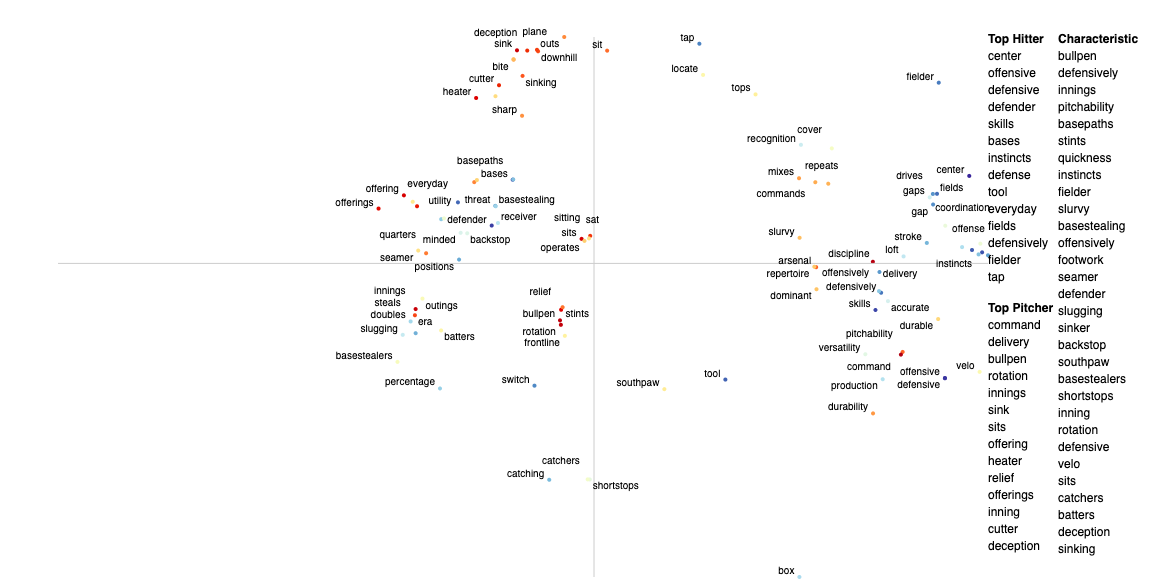}
\captionof{figure}{t-SNE visualization of reports by position.}

Before attempting classification, we can examine a subset of the corpus through visualization using the \texttt{Scattertext} python package \cite{Kessler_2017}. t-SNE is used to project word embeddings into a 2-dimensional space, leading to words used in similar contexts (similar reports) being closer together. Interestingly, several distinct clusters of language emerge; words like "relief," "bullpen", and "stints" bunch up with "rotation" and "frontline", indicating a set of contexts in pitchers' reports in which their roles are discussed. As well, words like "pitchability," "versatility", "command", "production", and "durability" form another region, as do "arsenal", "repertoire", "commands", and "mixes", which are likely contexts that discuss pitchers' abilities. As for hitters, a strong base-stealing cluster appears with "threat", "bases", "basepaths", and "basestealing" all bearing similarity. This exploratory analysis demonstrates the different ways and contexts in which scouts discuss players.

\section{Classification Task}

\subsection{Task Definition}

To demonstrate the utility of our dataset, we seek to answer the research question: "Using a player's scouting report, can we predict if they will make the major leagues?" We use several deep learning models implemented using the AllenNLP library \cite{Gardner2017AllenNLP}, which are made available on Github\footnote{\url{https://github.com/jacobdanovitch/jdnlp}}.

To prevent data leakage, it is crucial to mask certain tokens in the scouting reports such as player names, numeric quantities, and to a lesser extent, references to names of organizations. To illustrate the importance of this step, we conduct a study using the scaled F1 measure proposed in \cite{Kessler_2017}. We find that a text classification model could achieve 100\% recall by simply learning associations between mentions of the prospects in their scouting reports. As demonstrated in table \ref{tab:entitymask}, the most discriminative terms for positive classification (a successful prospect) are all names of prospects themselves, referenced in their own scouting reports. Thus, a model could infer by the presence of the term "Alford" that the report is about Blue Jays prospect Anthony Alford, who the model has previously seen the label for. To prevent this, we use the Natural Language Toolkit (NLTK) \cite{Loper:2002:NNL:1118108.1118117} to mask named entities with a special token. Numeric quantities can also have a similar effect - reports may refer to higher numeric values (throwing harder, a higher signing bonus, a high batting average), which could cause the same effect. We elect to mask this data as well, though the necessity of doing so is less than that of masking players' names.

\begin{table}[t]
\centering
\begin{tabular}{|l|l|l|}
\hline
\textbf{Term} & \textbf{Neg freq} & \textbf{Pos freq} \\ \hline
alford & 0 & 47 \\ \hline
nix & 0 & 43 \\ \hline
cecchini & 0 & 38 \\ \hline
robles & 0 & 35 \\ \hline
banda & 0 & 34 \\ \hline
fried & 0 & 31 \\ \hline
arroyo & 0 & 30 \\ \hline
ciuffo & 0 & 30 \\ \hline
tellez & 0 & 29 \\ \hline
grisham & 0 & 29 \\ \hline
\end{tabular}
\caption{The most discriminative terms present for the classification task.}
\label{tab:entitymask}
\end{table}

We apply labels using the age of 24 as a cutoff; if a player has not made their major-league debut by the time they turn 24, they are considered to have been unsuccessful as prospects. While this is certainly a harsh criterion, this task is primarily meant to demonstrate the potential applications of the dataset rather than to conduct a rigorous experiment. We exclude players who have not yet turned 24 and have not made their debuts, as they are too young to evaluate; however, we include players younger than 24 who \textit{have} made their debuts, as they have already demonstrated success. This coding scheme results in just over 7,000 usable data points with a heavy class imbalance (almost 80\% negative). This makes intuitive sense, as the majority of prospects do not become successful MLB players.

Due to the limited size of the dataset relative to most deep learning tasks, we perform data augmentation techniques such as word deletion and synonym substitution to double the size of the dataset. We also upsample positive examples, and use a weighted Cross-Entropy loss in each classifier (except for the Biattentive Classification Network, due to implementation constraints). All details are available with the model implementations at the aforementioned Github repository. 

\subsection{Results}

\begin{table}[h]
\centering
\begin{tabular}{|l|l|l|l|}
\hline
\textbf{Model} & \textbf{Accuracy} & \textbf{F-1 Score} \\ \hline
Bag Of Embeddings & 64.65\% & 53.78\% \\ \hline
TextCNN & 69.02 \% & \textbf{56.42}\% \\ \hline
LSTM+Self-Attention & 68.64\% & 54.65\% \\ \hline
Biattentive Classification Network & \textbf{73.52} \% & 43.33\% \\ \hline
Hierarchical Attention Network & 66.00\% & 54.07\% \\ \hline
\end{tabular}
\caption{Model performance for classifying success using scouting reports.}
\label{tab:clf}
\end{table}

Table \ref{tab:clf} demonstrates the performance of the various classifiers used as benchmarks. The loss-weighted classifiers show some degree of predictive accuracy while also maintaining reasonable F-! scores, as opposed to the BCN which quickly overfit to the majority class. Overall, the TextCNN demonstrated the best performance on the task relative to the two metrics. We hypothesize that this is due to the hierarchical, unordered structure of the documents. each sentence generally contains an important fact about the player, though the order in which these facts are presented is usually unimportant. CNNs are able to detect local n-grams within sentences, and pooling operations can identify which n-grams are discriminative \cite{Jacovi2018UnderstandingCN}. However, the improvement remains marginal at best, as all benchmarks perform quite similarly. Overall, these results provide moderate evidence for the possibility that scouts are able to author discriminative descriptions of players that can be used to predict the future performance of amateur baseball players.

\section{Interactive Application}

\includegraphics[scale=0.31]{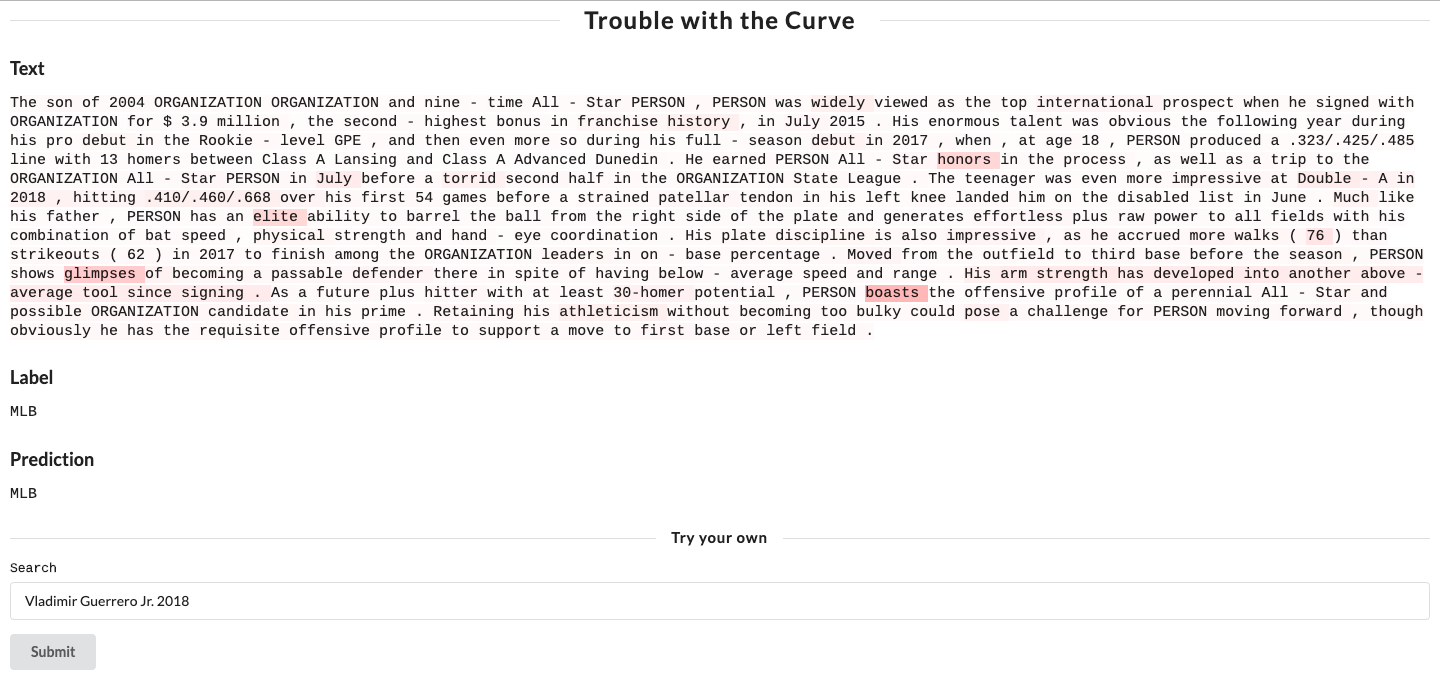}
\captionof{figure}{User interface to interact with a trained model.}

Lastly, we also present an interactive web application to visualize the outputs of the Hierarchical Attention Network \cite{Yang2016HierarchicalAN}. We chose this network due to its use of the attention mechanism, which provides interpretability towards it classifications. In the example above, the report of Vladimir Guerrero Jr., a highly-touted prospect, is classified positively. The model attends to generally complimentary words (in their respective contexts) like "honors", "elite", "glimpses", "above-average", and "boasts", demonstrating interesting insight toward the positive classification. The web application, as well as appropriate credit for the aesthetic portions of the user interface, accompany the aforementioned Github repository.

\section{Discussion}

In this work, we present the \textit{Trouble with the Curve} (TWTC) dataset of scouting reports for amateur baseball players. Each report contains written descriptions, numeric grades, and player identifiers. We then use this dataset to predict if an amateur baseball player will make the major leagues. We show that written descriptions provided by baseball scouts hold some predictive capacity for this task, providing an endorsement for the role of the scout in major league baseball. The dataset used for this task is shared with the community.

As seen in previous works, statistical approaches have also shown success in similar tasks. Future contributions could look to integrate player statistics with the written descriptions seen here to create a rich set of features, which we believe would greatly improve upon the results seen here. Other potential directions could include analyses of the scouting reports themselves, analyzing the language use in the written descriptions and evaluating the predictive capabilities of the numeric grades.

\bibliographystyle{plain}
\bibliography{references}

\end{document}